\documentclass[fleqn,10pt]{wlscirep}
\usepackage[utf8]{inputenc}
\usepackage[T1]{fontenc}
\usepackage{multirow}
\usepackage{array}
\usepackage{float}
\title{A multimodal Bayesian Network for symptom-level depression and anxiety prediction from voice and speech data}

\author[1]{Agnes Norbury}
\author[1]{George Fairs}
\author[1,2]{Alexandra L. Georgescu}
\author[3,4]{Matthew M. Nour}
\author[1]{Emilia Molimpakis}
\author[1*]{Stefano Goria}
\affil[1]{thymia Limited, London, UK}
\affil[2]{Institute of Psychiatry, Psychology \& Neuroscience, King’s College London, London, UK}
\affil[3]{Department of Psychiatry, University of Oxford, Oxford, UK}
\affil[4]{Max Planck UCL Centre for Computational Psychiatry and Ageing, University College London, London, UK}

\affil[*]{\texttt{stefano@thymia.ai}}

\begin{abstract}

During psychiatric assessment, clinicians observe not only what patients report, but important nonverbal signs such as tone, speech rate, fluency, responsiveness, and body language. Weighing and integrating these different information sources is a challenging task and a good candidate for support by intelligence-driven tools -- however this is yet to be realized in the clinic. Here, we argue that several important barriers to adoption can be addressed using Bayesian network modelling. To demonstrate this, we evaluate a model for depression and anxiety symptom prediction from voice and speech features in large-scale datasets (30,135 unique speakers). Alongside performance for conditions and symptoms (for depression, anxiety ROC-AUC=0.842,0.831 ECE=0.018,0.015; core individual symptom ROC-AUC>0.74), we assess demographic fairness and investigate integration across and redundancy between different input modality types. Clinical usefulness metrics and acceptability to mental health service users are explored. When provided with sufficiently rich and large-scale multimodal data streams and specified to represent common mental conditions at the symptom rather than disorder level, such models are a principled approach for building robust assessment support tools: providing clinically-relevant outputs in a transparent and explainable format that is directly amenable to expert clinical supervision.

\end{abstract}

\begin{document}

\flushbottom
\maketitle
%
%
\thispagestyle{empty}

\newpage

\section*{Introduction}

Psychiatric diagnosis represents one of medicine's most complex inferential challenges. During clinical assessment, psychiatrists must integrate multiple sources of information to arrive at diagnostic formulations that guide treatment decisions, in a process that amounts to inference to the best explanation given available evidence \cite{kind_how_2025,huda_medical_2021,aftab_psychiatric_2024}.

Psychiatry stands apart from other medical specialties in its reliance on subjective information sources. While neurologists integrate nerve conduction studies, brain imaging, and cerebrospinal fluid analysis with clinical presentation, psychiatric assessment depends primarily on clinical observation, patient self-report, and collateral information from family members or caregivers. The absence of established biological markers or objective diagnostic tests means that psychiatric diagnosis relies heavily on clinicians' abilities to synthesize complex, often ambiguous behavioural and self-report data \cite{bhugra_clinical_2011}.

The multimodal nature of psychiatric assessment reflects the complexity of mental health presentations. Experienced clinicians attend not only to reported symptoms, but to how these are communicated: including variations in tone of voice, speech rate, fluency, and responsiveness during conversational exchanges. Other important observations are body language, psychomotor activity, and cognitive processing as assessed through clinical interview and simple tasks \cite{kind_how_2025, casey_fishs_2019}. These paralinguistic and behavioural cues -- formalised in the “appearance and behaviour” sections of a typical mental state examination -- are particularly valuable given that mental health conditions often impact executive functioning and insight \cite{trzepacz_psychiatric_1993}.

Compounding this complexity, several factors influence the quality and consistency of assessment in a typical clinical setting. Clinicians often operate under significant time and capacity constraints \cite{royal_college_of_psychiatrists_workforce_2020,giotakos_psychiatry_2025}. The cognitive demands of gathering, weighing, and synthesizing multiple sources of uncertain information represent a substantial challenge even for experienced practitioners. Additionally, various forms of bias can affect which information is gathered, how it is weighted, and how diagnostic conclusions are reached -- which can lead to inequities in both access to assessment and diagnostic outcomes \cite{mouchabac_improving_2021,bansal_understanding_2022,clery_mental_2025}.

We propose that digital tools can augment psychiatric assessment by quantifying the paralinguistic, cognitive, and behavioural signals that clinicians observe but cannot systematically measure. Rather than replacing clinical judgment, these tools offer a form of "digital phenotyping" that makes explicit and measurable the characteristics that experienced clinicians intuitively recognise as diagnostically relevant. Such tools could bring psychiatry closer to the quantified precision medicine approaches available in other medical specialties, whilst also also supporting richer symptom monitoring beyond the clinical encounter. To date, intelligence-driven approaches have shown promise in predicting disorder status from voice, speech, movement, and physiological data -- however, their clinical translation remains limited \cite{chia_digital_2022}. Adoption of these approaches in the clinic has been hindered by a focus on binary condition classification rather than symptom- or sign-level assessment, a reliance on small sample sizes that compromises generalizability, poor transparency or explainability of underlying models, and lack of integration or congruence with existing clinical workflows \cite{huckvale_toward_2019,martin_estimating_2024,rutowski_toward_2024}.

Here, we argue that Bayesian network models with symptom-level estimation capacities are a promising approach for addressing these short-comings. These represent both a principled method for modelling the process by which clinicians must integrate across multiple noisy information sources during psychiatric assessment \cite{kyrimi_comprehensive_2021,polotskaya_bayesian_2024} and a natural way of accounting for the comorbidity and symptom co-occurrence patterns that are the norm rather than the exception in mental health \cite{mcgrath_comorbidity_2020}. From a clinical point of view, such models are able to deal well with heterogeneity within diagnostic categories \cite{forbes_elemental_2024}, provide the granular information required for treatment planning and progress monitoring \cite{waszczuk_what_2017,fried_depression_2015,fried_moving_2017,cross_tracking_2015}, and enable direct validation against clinical impressions and patient experience. From a modelling perspective, individual outputs are made more robust and explainable through integration of data-driven patterns with expert-derived knowledge, relative resilience to missing or noisy individual inputs, and transparent probabilistic reasoning. Crucially, Bayesian networks also support direct intervention in model predictions by supervising clinicians -- for example as more information becomes available about the context of specific symptoms during evaluation -- maintaining the primacy of expert judgement (from both clinicians and patients) during clinical decision-making \cite{dalfonso_ethical_2025}.

To demonstrate these advantages in practice, we present a novel Bayesian network-based assessment for common mental health conditions (depression and anxiety) and their associated symptoms. We describe a model built using multimodal voice and speech features -- although our approach would be naturally extendable to accommodate other relevant information sources such as cognitive test or passive-sensing data (see Discussion). Critically, this implementation leveraged multiple speech samples from over 30,000 unique participants, representing, to our knowledge, the largest dataset of its kind in psychiatric digital phenotyping research. This scale enabled application of rigorous best practices during both model development and evaluation. Specifically, sufficiently large samples were available for robust development and output calibration training, as well as independent test evaluation with appropriate numbers of observations across relevant demographic and clinical subgroups to support comprehensive fairness testing. This is vital for addressing both the documented brittleness of speech-based mental health prediction models trained on small, homogeneous samples \cite{berisha_are_2022,rutowski_toward_2024}, and concerns about generalizability and fairness in digital phenotyping tools more generally \cite{huckvale_toward_2019}. Alongside assessment of the true multimodal integration capacities of this model, we evaluate other characteristics of key translational relevance including clinical usefulness metrics and factors governing acceptability to mental health service users.

\section*{Results}

\subsection*{A multimodal Bayesian Network for depression and anxiety prediction from speech and voice data}

\begin{figure}[h]
    \makebox[\textwidth][c]
    {\includegraphics[width=1.05\textwidth]{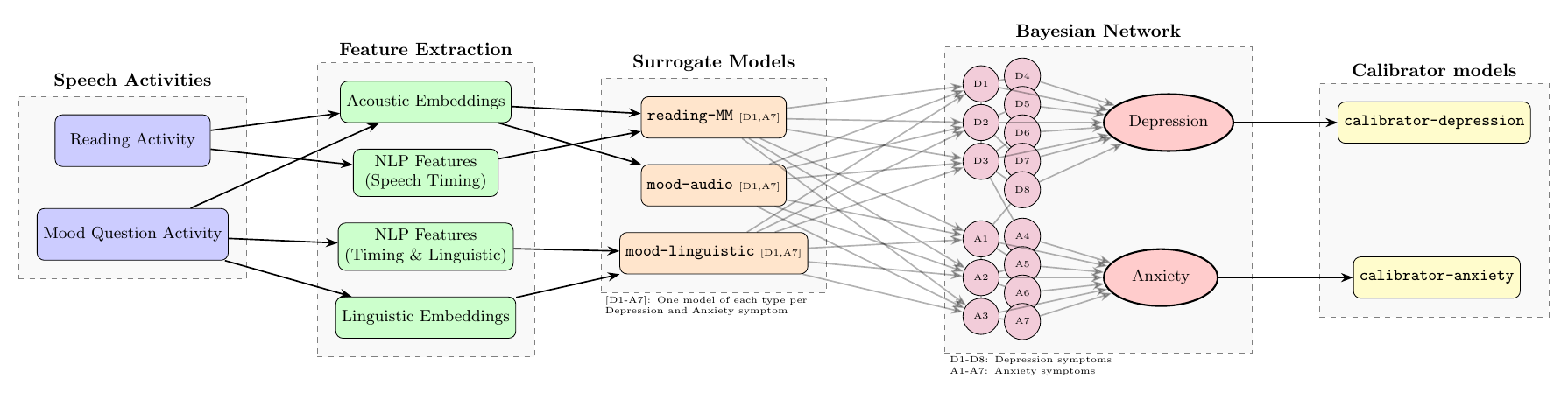}}
    \caption{\textbf{Model overview}. Speech activity data (reading out loud and answering a question about recent mood) is used to generate acoustic embeddings, speech timing, and linguistic feature sets (semantic embeddings and Natural Language Processing features), which are fed into relevant surrogate models to generate multiple predictions for each individual depression and anxiety symptom (for details, see \autoref{fig:surrogate-architecture}). Symptom-level predictions are passed to a Bayesian network, which specifies mapping weights of surrogate predictions to symptom severity estimates, inter-symptom relationships, and symptom severity to overall condition probabilities (simplified sketch of network architecture; for details see \autoref{fig:bn-structure}). Finally, condition probabilities are passed through a calibration layer to ensure meaningful output scores. } \label{fig:apollo-sketch}
\end{figure}

An overview of the model is provided in \autoref{fig:apollo-sketch}. Briefly, paralinguistic (acoustic and timing) and linguistic features were extracted from two different speech activities (reading out loud, and talking about recent mood). These were compressed into feature-type and symptom-specific representations using a set of surrogate models (neural networks trained to predict individual symptom presence from different subsets of paralinguistic and linguistic speech features). A joint Bayesian Network for depression and anxiety symptoms, alongside overall condition probabilities, was specified using insights from the clinical literature and parameterized using a dataset of surrogate model predictions, symptom and condition level ground truth values. This allowed the network to learn in parallel posterior probability distributions for the strength of relationships between different data types and true symptom severity scores, relationships between different symptoms, and relationships between symptom severity states and overall condition probabilities. In a final step, calibrator models were trained to ensure that output condition probability scores were well-aligned to observed case frequencies for depression and anxiety, respectively.

Development data for surrogate models and Bayesian Network training and evaluation were available from \textit{N}=21,379 participants. A separate calibration set from \textit{N}=6,325 participants was available for training of the calibrator models. Final performance evaluation was carried out in a held-out test set (\textit{N}=2,431 participants), which was completely unseen during model training and development. For a description of participants in each dataset, please see \autoref{tab:demogs}.

\subsubsection*{Model performance for conditions}

Discrimination and calibration performance for overall condition status (depression and anxiety) are described in \autoref{tab:condition-performance}. These two metrics assess important independent model qualities: that the model is able to distinguish well between users with different condition statuses, and that output probabilities are meaningful (i.e., that a 75\% probability score actually corresponds to 75\% positive condition rate) \cite{hond_interpreting_2022,calster_performance_2024}. Performance results in the held-out test should be interpreted as representing generalized estimates of model performance, with results from the development set test split (data unseen during surrogate model training but available during Bayesian Network development) provided for reference and as an indicator of performance stability. 

\begin{table}[h]
\centering
\begin{tabular}{@{}>{\raggedright\arraybackslash}p{2cm}>{\raggedright\arraybackslash}p{2cm}>{\centering\arraybackslash}p{2.5cm}>{\centering\arraybackslash}p{2.5cm}@{}}
\toprule
\multicolumn{1}{@{}l}{} & \multicolumn{1}{l}{} & \multicolumn{1}{c}{\begin{tabular}[t]{@{}c@{}}Development\\ (Uncalibrated)\end{tabular}} & \multicolumn{1}{c}{\begin{tabular}[t]{@{}c@{}}Test\\ (Calibrated)\end{tabular}} \\ \midrule
Depression & ROC-AUC & 0.837 & 0.842 \\
 & ECE & 0.060 & 0.018 \\
Anxiety & ROC-AUC & 0.836 & 0.831 \\
 & ECE & 0.090 & 0.015 \\ \bottomrule 
\end{tabular}
\caption{\textbf{Model discrimination and calibration performance for overall condition probabilities for depression and anxiety}. The development set test split (depression, anxiety \textit{N}=2251, 2191) was unseen during surrogate model training but used for development of the final Bayesian Network model architecture. The held-out test set (depression, anxiety \textit{N}=1489, 1477) was completely unseen during model training and development, with disorder probabilities calibrated using calibrator models trained on a further held-out calibration set. \textit{ROC-AUC}, Area Under the Receiver-Operator Curve; \textit{ECE}, Expected Calibration Error.}
\label{tab:condition-performance}
\end{table}

\begin{figure}[h]
    \makebox[\textwidth][c]
    {\includegraphics[width=0.8\textwidth]{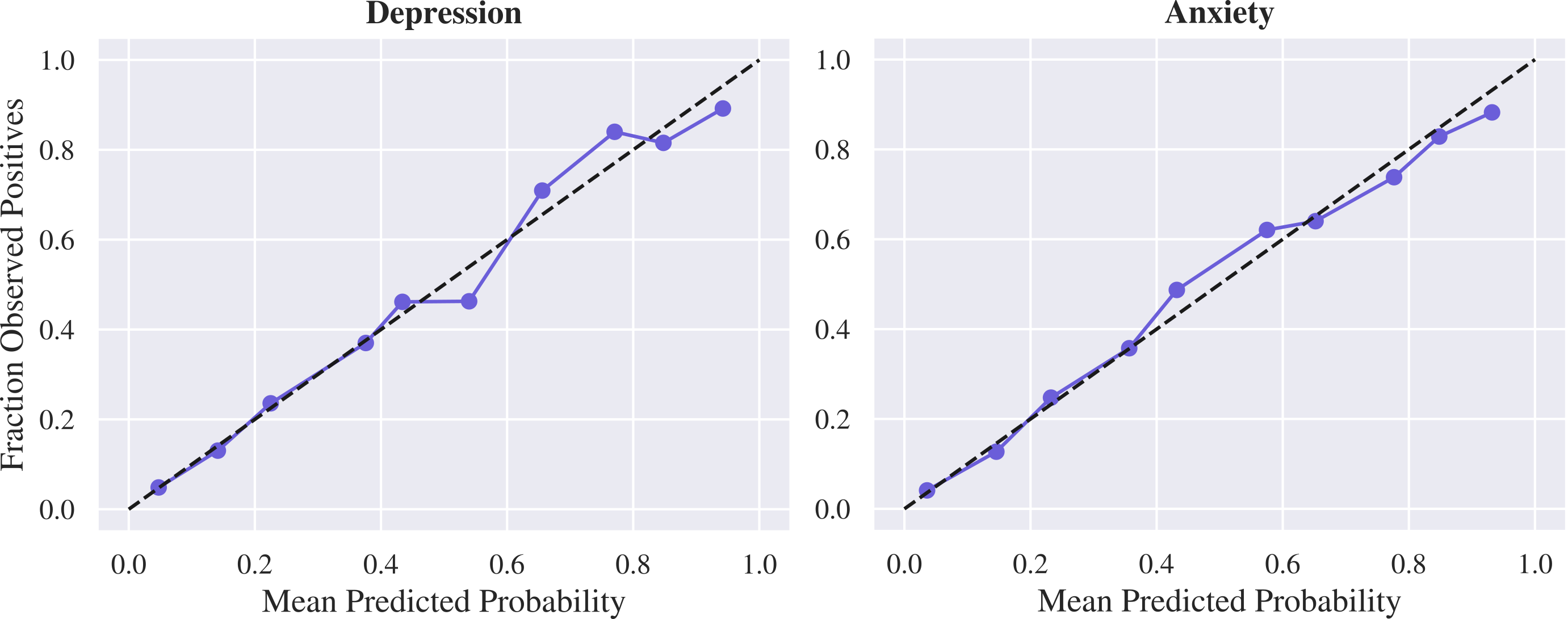}}
    \caption{\textbf{Model calibration for overall depression and anxiety and status}. Plots represent predicted probability score ranges \textit{vs} observed positive case rate values across the full range of output condition probabilities in the held-out test set. The dotted line at $y=x$ represents performance of a perfectly calibrated model.} \label{fig:calib-curves}
\end{figure}

Although specific benchmarks for interpreting discrimination and calibration performance measures are somewhat arbitrary and inconsistently applied in the literature, as a general rule-of-thumb ROC-AUC values above 0.80 and ECE values below 0.05 are often considered to be properties of clinically-useful prediction models \cite{hond_interpreting_2022, guo_calibration_2017}. Full calibration curve plots for condition probabilities (based on calibrated predictions in the held-out test set) are shown in \autoref{fig:calib-curves}.

\subsubsection*{Model performance for individual symptoms}

Discrimination performance for individual depression and anxiety symptoms is described in \autoref{tab:symptom-performance}. These metrics quantify the ability of the model to discriminate between cases of significant symptom presence \textit{vs} absence (based on the threshold of symptom experience on "half or more days" \textit{vs} "less than half of days" over the last 2 weeks, as described in DSM criteria for depressive and anxiety disorders, and applied in previously-published algorithms for converting combined symptom severity/frequency ratings on self-reported questionnaires to DSM-like criteria \cite{kroenke_phq-9_2001}).

Classification performance for individual symptoms was generally in the fair-to-good range ($>0.70$ ROC-AUC \cite{hond_interpreting_2022}), with stronger performance ($\sim0.75$ ROC-AUC) observed for "core" symptoms (anhedonia, low mood, anxiety/nervousness, and uncontrollable worry). There was evidence that some symptoms may may benefit from augmentation of the network with non-speech data sources: for example cognitive task data may provide additional information about concentration problems and restlessness symptoms (see Discussion).

\begin{table}[h]
\centering
\resizebox{\textwidth}{!}{%
\begin{tabular}{@{}rcccccccc@{}}
\toprule
\multicolumn{9}{c}{Depression Symptoms} \\ \midrule
\multicolumn{1}{l}{} & Anhedonia & Low Mood & Sleep & Low Energy & Appetite & Worthlessness & Concentration & Psychomotor \\
Development & 0.732 & 0.785 & 0.709 & 0.734 & 0.697 & 0.761 & 0.696 & 0.721 \\
Test & 0.741 & 0.801 & 0.706 & 0.734 & 0.690 & 0.757 & 0.667 & 0.714 \\ \midrule
\multicolumn{9}{c}{Anxiety Symptoms} \\ \midrule
\multicolumn{1}{l}{} & Nervousness & Uncontrollable Worry & Excessive Worry & Trouble Relaxing & Restlessness & Irritability & Dread \\
Development & 0.761 & 0.751 & 0.747 & 0.724 & 0.696 & 0.718 & 0.705  \\
Test & 0.749 & 0.746 & 0.736 & 0.739 & 0.680 & 0.706 & 0.721  \\ \bottomrule
\end{tabular}%
}
\caption{\textbf{Model performance for individual depression and anxiety symptoms}. Values represent ROC-AUC scores for symptom presence (for half of days or more) \textit{vs} absence (less than half of days), in the development set test split (\textit{N}=3372, 3378 for depression, anxiety symptoms) and the fully held-out test set (\textit{N}=2426, 2424 for depression, anxiety symptoms).}
\label{tab:symptom-performance}
\end{table}

\subsubsection*{Dealing with clinical heterogeneity}

Given the heterogeneity of common mental health conditions in terms of observed symptom profiles (particularly for depression \cite{forbes_elemental_2024}), we examined robustness of discrimination performance to different condition definitions and presentation types (where possible, using previously-established algorithms for defining these from self-report data).

For DSM-like Major Depressive Disorder, defined as 5 or more depressive symptoms having been present at least "more than half the days" in the past 2 weeks, 1 of which must be depressed mood or anhedonia\cite{kroenke_phq-9_2001}), ROC-AUC values were 0.836, 0.852 in the development and test sets, respectively. 

For DSM-like Other Depression, defined as 4 or more depressive symptoms having been present at least more than half the days in the past 2 weeks\cite{kroenke_phq-9_2001}, ROC-AUCs were 0.814, 0.829, respectively.

For DSM-like Generalized Anxiety Disorder, no previously-published algorithm exists, so this was derived from DSM-5 criteria (which specify 5 or more anxiety symptoms having been present at least "more than half the days" in the past 2 weeks, 1 of which must be feeling nervous, anxious or edge or uncontrollable worry), ROC-AUCs were 0.819, 0.835, respectively.

This reveals that, whilst performance is slightly superior for typical depression presentations (involving significant alterations in mood and motivation levels), it remains within strong ranges for less typical presentations (e.g., those more dominated by somatic changes). In general, a model which is able to maximise performance across different possible individual symptoms should be better placed to deal with heterogenous presentations than one which simply predicts overall condition absence or presence according to sum severity cutoff values. 

\subsubsection*{Condition severity}

The Bayesian network analysed here represents symptoms not just in terms of binary presence or absence, but specifies posterior probability distributions across 4 possible severity categories. These can be used to calculate the expected severity level for each symptom, which can then be summed across conditions to output overall disorder severity predictions alongside condition probabilities. To validate these severity estimates, predicted values were compared to observed PHQ-8 and GAD-7 total scores (which similarly represent overall symptom load based on number of observed symptoms and their severity). Condition severity predictions were found to be strongly associated \cite{hemphill_interpreting_2003} with these scores: for depression, Pearson's $\textit{r}=0.526, 0.551$; for anxiety, $\textit{r}=0.514, 0.513$ in the development test split, and held out test set, respectively.

We also assessed whether severity estimates were related to other important patient-reported outcomes, in particular the impact of experience of mental health symptoms on quality of life and psychosocial functioning. Mental-Health related Quality of Life \cite{van_krugten_mental_2022} scores were available in a subset of development set test split participants (\textit{N}=804), and observed to be correlated with predicted depression severity at $\textit{r}=-0.49$ and with anxiety severity at $\textit{r}=-0.50$. In the held-out test set (\textit{N}=2413), PHQ psychosocial functioning (level of impairment in work, home, and social life\cite{kroenke_phq-9_2001}) and the CDC's healthy days (number of recent days where poor physical or mental health limited usual activities\cite{moriarty_centers_2003}) measures were available: and were related to model-predicted depression and anxiety severity at $\textit{r}=0.47, 0.47$ and $\textit{r}=0.44, 0.42$, respectively.

\subsubsection*{Fairness}

A key requirement for digital diagnostic support tools is to ameliorate (and not exacerbate) existing structural inequalities in diagnostic practices and healthcare access \cite{nice_digital_2025}. We therefore robustly assessed model performance for potential differences in performance for members of different demographic groups. Group-level discriminative performance (ROC-AUC values) and between-group differences in calibration metrics (ECE), combined calibration-accuracy measures (Brier scores), and outcome fairness (equalized odds ratios, which assess potential differences in true positive, true negative, false positive, and false negative rates between groups) are reported in \autoref{tab:fairness}.

\begin{table}[h]
\resizebox{\textwidth}{!}{%
\begin{tabular}{rrcccccccccc}
\hline
\multicolumn{1}{l}{} & \multicolumn{1}{l}{} & \multicolumn{5}{c}{Depression} & \multicolumn{5}{c}{Anxiety} \\ \hline
\multicolumn{1}{l}{} & Group & \textit{N} & \begin{tabular}[c]{@{}c@{}}ROC-\\ AUC\end{tabular} & \begin{tabular}[c]{@{}c@{}}ECE\\ difference\end{tabular} & \begin{tabular}[c]{@{}c@{}}Brier Score\\Difference\end{tabular} & \begin{tabular}[c]{@{}c@{}}Equalized Odds\\Ratio Difference\end{tabular} & \textit{N} & \begin{tabular}[c]{@{}c@{}}ROC-\\ AUC\end{tabular} & \begin{tabular}[c]{@{}c@{}}ECE\\Difference\end{tabular} & \begin{tabular}[c]{@{}c@{}}Brier Score\\Difference\end{tabular} & \begin{tabular}[c]{@{}c@{}}Equalized Odds\\Ratio Difference\end{tabular} \\ \hline
Age & <35 & 720 & 0.826 & 0.006 & 0.033 & 0.099 & 705 & 0.814 & 0.002 & 0.042 & 0.047 \\
 & $\geq35$ & 769 & 0.848 &  &  &  & 772 & 0.836 &  &  &  \\
Birth Sex & Female & 974 & 0.838 & 0.036 & 0.037 & 0.111 & 972 & 0.825 & 0.017 & 0.044 & 0.251 \\
 & Male & 515 & 0.828 &  &  &  & 505 & 0.800 &  &  &  \\
Gender Identity & Woman, Non-binary & 975 & 0.834 & 0.038 & 0.043 & 0.067 & 974 & 0.820 & 0.022 & 0.052 & 0.177 \\
 & Man & 514 & 0.842 &  &  &  & 503 & 0.812 &  &  &  \\
Race/Ethnicity & White & 1086 & 0.850 & 0.055 & 0.009 & 0.024 & 1068 & 0.843 & 0.061 & 0.005 & 0.060 \\
 & Other & 404 & 0.840 &  &  &  & 409 & 0.814 &  &  &  \\
Accent & UK & 617 & 0.835 & 0.016 & 0.018 & 0.085 & 609 & 0.821 & 0.015 & 0.030 & 0.167 \\
 & US, Other & 862 & 0.846 &  &  &  & 860 & 0.831 &  &  &  \\
Chronic Health Condition & No & 894 & 0.839 & 0.029 & 0.050 & 0.038 & 887 & 0.835 & 0.024 & 0.054 & 0.050 \\
 & Yes & 595 & 0.840 &  &  &  & 590 & 0.814 &  &  &  \\
Device Type & Laptop & 876 & 0.831 & 0.006 & 0.008 & 0.054 & 880 & 0.839 & 0.003 & 0.028 & 0.106 \\
 & Smartphone, Tablet & 607 & \multicolumn{1}{l}{0.856} &  &  &  & 591 & 0.819 &  &  &  \\ \hline
\end{tabular}%
}
\caption{\textbf{Differences in discrimination and calibration performance for anxiety and depression across different demographic groups}. Results are for calibrated condition probabilities in the test set. \textit{ROC-AUC}, single group discrimination performance for each condition; \textit{ECE Difference}, difference in Expected Calibration Error between groups; \textit{Brier Score Difference}, difference in Brier scores (a combined discrimination and calibration metric that measures accuracy of probabilistic predictions) between groups; \textit{Equalized Odds Ratio Difference}, difference in a combined measure of true positive, true negative, false positive, and false negative rates between groups. A $0.05$ difference in Brier score means predictions differ by approximately $\sqrt{0.05}=0.22$ on a probability scale; equalized odds ratio differences can be directed interpreted on a probability scale.}
\label{tab:fairness}
\end{table}

Within all examined subgroups, ROC-AUC for both depression and anxiety remained $\geq0.80$, indicating maintenance of clinically-appropriate performance levels for specific groups. This also implies that model performance is unlikely to be significantly driven by between-group differences in voice or speech features, which in combination with differential rates of experience of common mental health problems between some groups has previously been shown to inflate performance estimates for some speech-based models \cite{goria_revealing_2024}.

There was some evidence of differences in probability calibration (Brier scores) between birth sex, gender and chronic health condition groups -- although these differences were all within the approximately 20\% range often used as a rule-of-thumb for fairness assessment (although this is somewhat a blunt heuristic which should take into account the context in which a particular tool will be used \cite{watkins_four-fifths_2022}). These could be addressed in future implementation using demographic-group specific calibration strategies. Minimal differences in calibration accuracy were observed for other tested groups including race/ethnicity, accent, and testing device type.

Importantly, differences in calibration properties were not in most cases associated with strong differences in outcome fairness, as assessed by differences in equalized odds ratios. This measures a model's risk for misclassifcation bias, or that members of different groups will experience differential allocation of treatment outcomes. For this metric, most group differences were in excellent ($<0.05$) or good ($<0.10$) ranges (representing <5\% or <10\% differences between groups), with only one tested difference exceeding the 20\% threshold. Differences in equalized odds ratios for anxiety between birth sex groups were found to be driven by higher true positive rates for females than males, which may be related to substantially higher base rates of anxiety for women \textit{vs} men in our training data. This difference warrants further attention and may benefit from mitigation strategies such as sex-based stratification of network training data and/or sex-specific calibration strategies for anxiety. Of note, equalized odds ratio measures are dependent on the specific threshold used for classification of cases \textit{vs} non-cases in the analysis, which can also be adjusted depending on required use-case characteristics.

\subsection*{Model properties}

\subsubsection*{Relationship between conditions}

Depression and anxiety often co-occur. For example, lifetime prevalence estimates for proportion of people with a depressive disorder who also meet diagnostic criteria for an anxiety disorder range from 49\% to 81\%; with similar rates for meeting depressive disorder criteria in people with an anxiety diagnosis (47\% to 88\%)  \cite{mcgrath_comorbidity_2020,jacobson_anxiety_2017}.

A joint model of depression and anxiety should take into account this comorbidity, whilst also allowing for condition-specific sensitivity in predictions (cases where anxiety is present but not depression, and vice versa). When marginalizing over all other model states, the posterior conditional probability for the presence of anxiety, given presence of depression in our network was 0.409, and the probability for the presence of depression, given the presence of anxiety, was 0.424. This degree of separation was influenced by cross-correlation of condition states in our training data, given the presence of regularizing Bayesian priors (see Methods). A strength of the Bayesian Network framework is that it allows for enforcement of different conditional probability estimates between conditions: either by applying different prior settings prior to training, or intervening directly on posterior estimates, if this is desired for different populations or implementation settings.

\subsubsection*{Multimodal integration at the symptom level}

We have argued that Bayesian Networks represent a principled method for integrating information about mental health symptoms across different data sources: by identifying which inputs carry greater signal for different symptoms, and integrating across these in an appropriate way to construct more accurate output estimates. Examining characteristics of our network inputs and inspecting the conditional posterior probability distributions it constructs from these during training allows us to see how well this works in practice.

\begin{figure}[h]
    \makebox[\textwidth][c]
    {\includegraphics[width=0.9\textwidth]{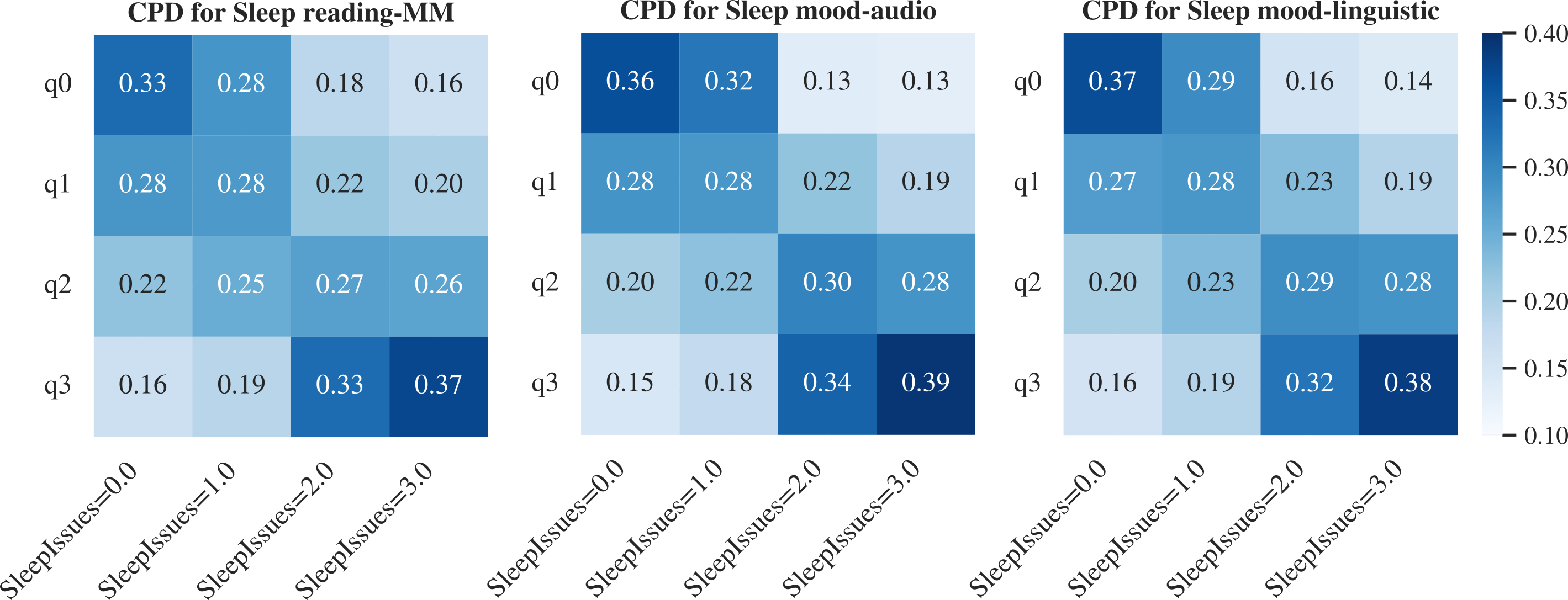}}
    \caption{\textbf{Multimodal integration}. Example posterior Conditional Probability Distributions (CPDs) for sleep symptom severity states (\textit{SleepIssues=0-3}) for different surrogate model types (\textit{q0}=lowest quartile, \textit{q3}=highest quartile of predicted symptom probability categories for each model). } \label{fig:example-CPDs}
\end{figure}

Different surrogate model performance for individual depression and anxiety symptoms is described in \autoref{tab:surrogate-symptom-rocs}. For every symptom, network performance is superior to that of any individual surrogate: suggesting the model successfully arbitrates between and integrates across the multiple noisy observable inputs it receives. For example, for sleep problems, surrogate model predictions with discriminative performances of 0.62, 0.66, and 0.68 (\texttt{reading-MM}, \texttt{mood-audio}, and \texttt{mood-linguistic}, respectively) are weighted and combined by the network to produce a final performance of 0.71 ROC-AUC (\autoref{tab:symptom-performance}). \autoref{fig:example-CPDs} shows that the network assigns different posterior probability weights to different sleep symptom severity states for each surrogate model input: with higher posterior probabilities assigned to more discriminative surrogates for that symptom. 

\subsubsection*{Relative contribution and redundancy between different input types}
\autoref{tab:single-surrogate-bn-performance} shows performance of the Bayesian Network when queried only with evidence from each surrogate model type (reading activity paralinguistic feature, mood activity paralinguistic feature, and mood activity linguistic feature predictions). Discriminative performance (ROC-AUC) for conditions was superior for \texttt{mood-audio} and \texttt{mood-linguistic} model predictions, compared to \texttt{reading-MM} predictions alone (although all were inferior to querying with the full set of surrogate types). Raw calibration performance (ECE) was superior when providing either paralinguistic compared to linguistic only model predictions.

It is important to note that the mood activity linguistic features may contain a natural semantic overlap with some PHQ/GAD items, as during this activity participants were explicitly asked to describe their mood over the past two weeks. Despite this advantage, we observed that providing the network with predictions from both paralinguistic surrogate types (\texttt{reading-MM} and \texttt{mood-audio}) resulted in matched discriminative performance, and improved calibration performance, to \texttt{mood-linguistic} predictions only for depression (both ROC-AUC$=0.783$; for anxiety ROC-AUC was slightly inferior at $0.783$ \textit{vs} $0.805$).

This indicates evidence of built-in redundancy -- or that good signal is available from multiple input data types (here, both paralinguistic and linguistic speech features). This is a desirable model property, as it helps build robustness to noise or fragility in any individual data type. For example, whilst linguistic-only models can show strong average discriminative performance for depression status when elicited semantic content directly relates to indicative symptoms, they are vulnerable to failure if particular individuals choose or neglect to disclose relevant semantic cues. Of note, our chosen network architecture enforced a design constraint whereby every symptom must be informed by at least one paralinguistic input, preventing over-reliance on linguistic features that may be fragile to individual disclosure patterns, cultural differences in expression, or vocabulary choices.
Given a key feature of Bayesian Network models is resilience to missing information, this redundancy also allows the model to still produce meaningful outputs (albeit at slightly degraded performance levels) if, for example a user or clinician decides that a particular input activity type should not be used for inference.

\subsubsection*{Intervening in model predictions: clinician-in-the-loop}

\begin{figure}[h]
    \makebox[\textwidth][c]
    {\includegraphics[width=0.95\textwidth]{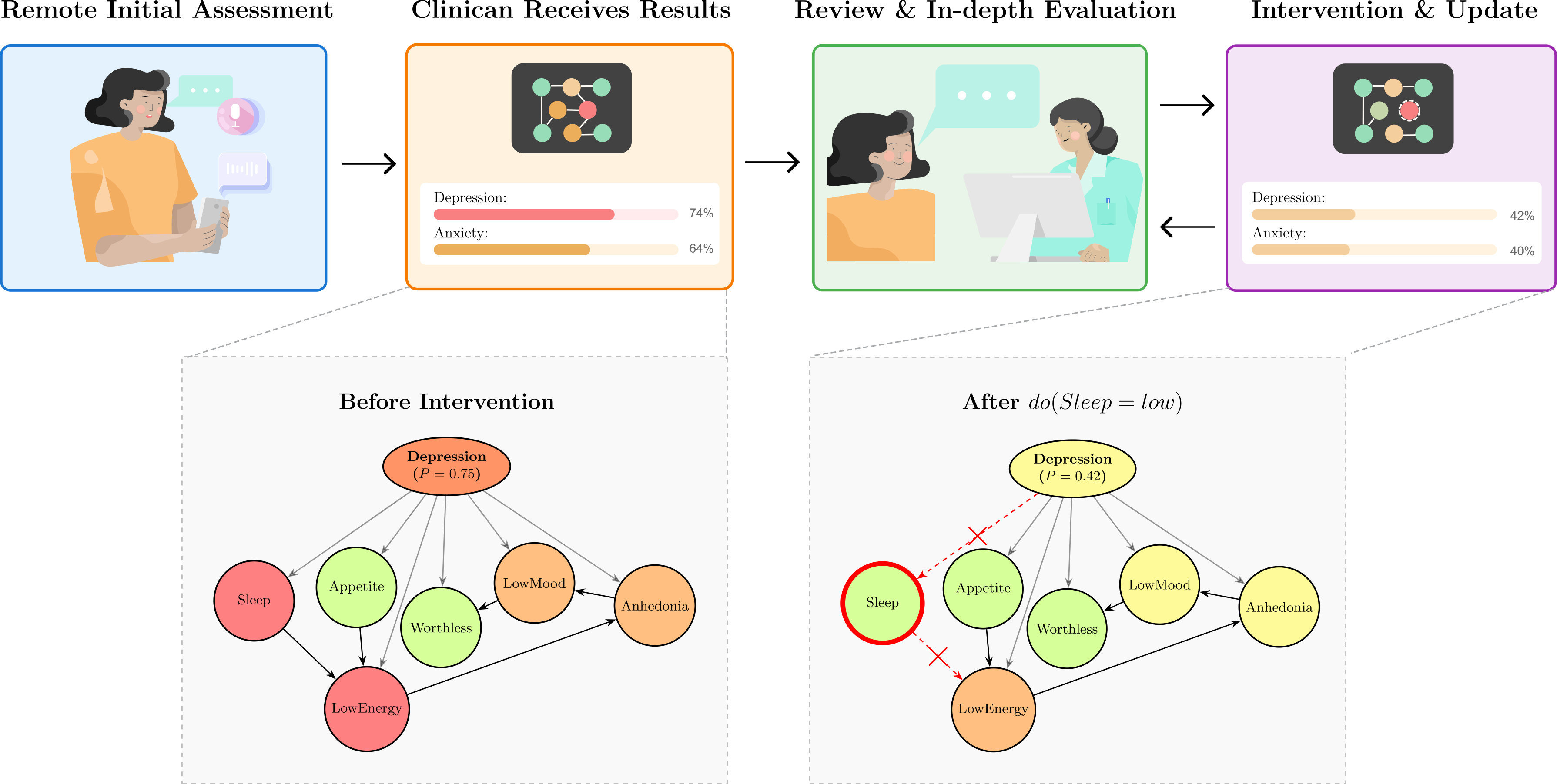}}
    \caption{\textbf{Intervening on network predictions}. Example of direct clinician intervention in model predictions based on follow-up discussions with a patient or client, using do-operations. For accompanying vignette, please see main text. Insets show a toy example for a subset of Bayesian Network depression symptom nodes, illustrating the effect of isolating sleep symptom predictions from the network after this has been evaluated as better explained by contextual rather than mental health-related factors.}\label{fig:example-intervention}
\end{figure}

Other key properties afforded by the Bayesian Network architecture are explainability: or explicit reporting of the contribution of individual symptom severity estimates to overall condition probabilities, and modifiability: or the ability of a supervising clinician to intervene directly in model predictions on the basis of follow-up discussions with their client. A vignette of this iterative update process, which allows the clinician and client to collaboratively update predictions based on more in-depth evaluation of individual problem areas is described below and in \autoref{fig:example-intervention}.

\textbf{Intervention by do-operation vignette}. Sleep disturbances and low energy are common in people experiencing both depression and anxiety, and are therefore often influential symptoms in condition network models. However, they can also be related to external life factors that in some cases may mean either clients or consulting clinicians do not consider it appropriate to evaluate them as mental health-related symptoms. For example, a user may complete an initial screening assessment which indicates moderate to high probabilities of depression and anxiety. On reviewing the results, the clinician is able to see that this is mainly driven by high severity estimates for sleep and fatigue symptoms. When discussing this result with the client, they discover that they have been sleeping very poorly over the past week or so due to caring for an unwell dependent. The clinician is able to incorporate this information into model predictions by isolating the sleep symptom node from the rest of the network (in Bayesian Networks, this can be achieved straightforwardly using a causal inference method known as a do-operation). This removes the influence of this symptom in an updated set of symptom-level and overall condition probabilities.

\subsection*{Clinical usefulness and acceptability}

An assessment support tool of the type described here will only useful if it can both provide actionable information and is acceptable to end-users. 

\subsubsection*{Screening tool usefulness: prevalence-dependent performance metrics}

In order to assess the potential usefulness of model outputs in a real-world clinical setting, we supplemented our previous performance analyses by calculating prevalence-dependent metrics for calibrated condition predictions in the held-out test set. The case rates for anxiety and depression in our test data were $\sim30\%$: a reasonable approximation for primary care settings in the UK, where it has been estimated that 4 in 10 consultations involve a mental health element \cite{mind_gp_2018}. For calculating prevalence-dependent metrics analysis, we used a threshold of 0.5 for classifying predicted probabilities as positive \textit{vs} negative cases -- different thresholds may be appropriate for different use cases, based on the relative costs and benefits of false positive \textit{vs} false negative results.

\begin{table}[h]
\centering
\label{tab:prevalence-metrics}
\begin{tabular}{@{}lllll@{}}
\toprule
 & PPV & NPV & LR+ & LR- \\ \midrule
Depression & 0.69 & 0.83 & 4.7 & 0.46 \\
Anxiety & 0.71 & 0.80 & 5.4 & 0.55 \\ \bottomrule
\end{tabular}
\caption{\textbf{Prevalence-dependent performance metrics for overall condition predictions}. \textit{PPV}, Positive Predictive Value (probability that a person from this population with a positive test result has a condition); \textit{NPV}, Negative Predictive Value (probability that a person from this population with a negative test result does not have the condition); \textit{LR+}, Positive Likelihood Ratio (change in odds of having a condition after receiving positive test result); \textit{LR-} Negative Likelihood Ratio (change in odds of having a condition after receiving a negative test result). }
\end{table}

Positive predictive value (PPV) is the proportion of true positives of all predictive positive cases. A PPV of $\sim0.70$ suggests that , at a threshold of 0.5, the model produces some false positive results. However, this is usually considered acceptable for initial risk screening or population monitoring applications, where the cost of missing cases typically outweighs that of false alarms. Higher negative predictive value (NPV) scores suggest that the model is reliable at ruling out conditions when the ground truth is negative. Positive Likelihood Ratio (LR+) scores of $\sim5$ can be considered strong: meaning that a user is $5x$ more likely to have a condition after receiving a positive test result. Negative Likelihood Ratios (LR-) are moderate: with scores of $\sim0.5$ implying half the chance of having a condition after a negative test result. This means that a negative test result is not definitive in ruling out a condition (at least, in a population with a relatively high baseline prevalence for common mental health conditions). 

Overall, these scores suggest good properties for an initial screening, triaging, or population-monitoring tool: with positive results requiring clinical follow-up for confirmation. Provision of well-calibrated condition probability scores allows stakeholders to select thresholds for further investigation most appropriate for their particular context (for example, via decision curve analysis).

\subsubsection*{Service user consultation and comparison to care as usual}

Based on previous descriptions of the experiences of clinicians and clients with standard monitoring approaches embedded in the UK NHS Talking Therapies program \cite{malpass_usefulness_2016,bendall_contending_2020,ford_use_2020,faija_using_2022}, we asked a sample of participants with lived or living experience of both common mental health problems and treatment within the UK health system to provide feedback on a proposed speech-based tool for screening and monitoring of depression and anxiety symptoms, in comparison to care-as-usual with standard questionnaires.

Survey participants (\textit{N}=230) are described in \autoref{tab:survey-demogs}. All participants had a diagnosis of a mental health condition (the most common being anxiety and depressive disorders), and had tried to access mental health support via the UK public health system (NHS) in the past 12 months. 96\% of participants reported that they were somewhat or very familiar with mental health screening and monitoring via standardized questionnaires.

\begin{figure}[H]
    \makebox[\textwidth][c]
    {\includegraphics[width=1.0\textwidth]{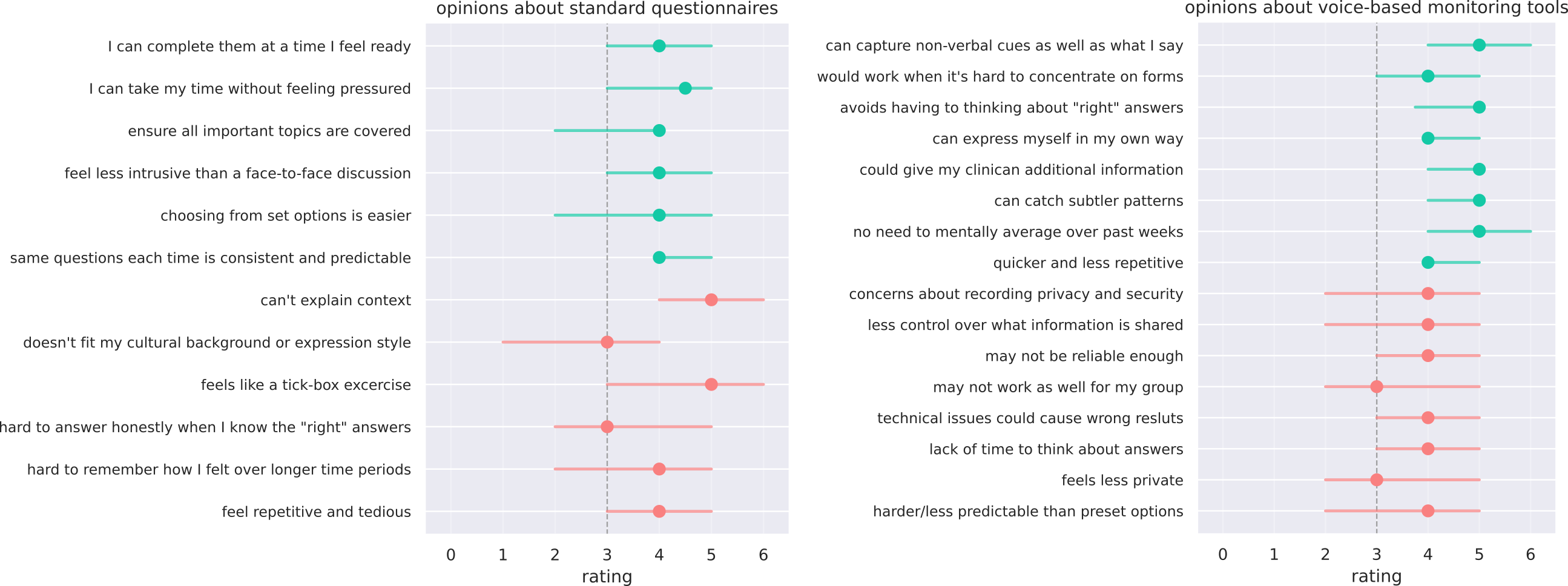}}
    \caption{\textbf{Mental health service user views of strengths and weaknesses of standard monitoring questionnaires and a hypothetical speech-based assessment tool}. Summary of responses from \textit{N}=230 participants with a mental health diagnosis and experience of screening and monitoring practices in UK mental health services. For each statement, survey participants were asked to indicated how much they agreed or disagreed on a 7-point likert scale (0-completely disagree, 6-completely agree). Plotted values represent interquartile ranges and medians across participant ratings.} \label{fig:questionnaires-voice-opinions}
\end{figure}

On average, survey respondents endorsed (agreed with) a range of opinions regarding strengths and weaknesses of the different assessment measure types (\autoref{fig:questionnaires-voice-opinions}). For standardized questionnaires, the most strongly endorsed strengths were \textit{being able to take time filling them out (without feeling pressured)}, and the \textit{predictability of content}. Consistent with previous qualitative research in UK mental health service users, the most commonly endorsed weaknesses were being \textit{unable to add nuance or explain the context behind answers}, and \textit{that completing them feeling like an impersonal tick-box exercise}. Participants who were very familiar with standard screening measures (filling them out regularly) showed a tendency towards more negative overall views of them, compared to participants who had only completed them a few times (Mann-Whitney U=6980, \textit{p}=0.05, rank-biserial correlation effect size=-0.146 on a 7-point preference scale). For voice-based screening tools (such as the one described here), the most strongly endorsed strengths were \textit{being able to capture non-verbal as well as semantic state information}, and \textit{not needing to mentally average the intensity and frequency of symptoms over the past weeks} into a single rating number. More commonly endorsed concerns were \textit{tool reliability}, the \textit{possibility of technical issues affecting results}, and \textit{having enough time to give considered spoken responses}. In this sample, average endorsement of potential strengths was numerically higher than for these concerns.

Overall, initial responses to multimodal voice and speech-based symptom monitoring approaches were positive (71\% of respondents were excited by or somewhat interested in the idea), with significant proportions of respondents either unsure (19\%) or somewhat or very concerned about the idea (6 and 4\%, respectively). Qualitative analysis of themes identified in open-ended feedback regarding features required for trustworthy implementation of voice-based mental health screening technologies is available in \autoref{tab:voice-tool-trustworthiness}. The most commonly raised themes were privacy, security, and control over use of speech recording data; provision of sufficient evidence of tool accuracy and reliability, including for diverse demographic groups; and endorsement by healthcare professionals or public health bodies. 7\% of participants reported a general reluctance to use any form of AI-based screening tool, on the basis of their current experience with such technologies.

These results provide a limited view of overall appetite for and concerns around implementation of speech and voice-based screening technologies, given that they are from a small, self-selected sample of research participation platform users, which was predominantly White and generally comfortable with technology use (see \autoref{tab:survey-demogs}). However, they do help identify some important considerations for any potential implementation of these methods in clinical practice. Specifically, that key to the realization of the potential benefits of these methods are that users retain control over how their voice data is stored and used, and how and when it is shared with their clinical team, that they should supplement rather than displace clinical contact time, and that traditional options remain available as an alternative for users with legitimate concerns that such methods will work for them \cite{nice_digital_2025}.

\section*{Discussion}

Here, we evaluated a novel approach to supporting assessment for common mental health problems, using a Bayesian network model built from multimodal voice and speech features to explicitly represent individual depression and anxiety symptoms. Analysis of model properties showed that the network was successful in arbitrating between and integrating across multiple noisy input data sources for each symptom -- resulting in excellent discrimination and calibration properties for both conditions, and good performance for the majority of individual symptoms. Providing symptom-level outputs is crucial for clinical adoption of assessment support tools in mental health care, as symptoms, not diagnoses often guide clinical practice in psychiatry \cite{waszczuk_what_2017}. As symptoms both occur across diagnostic boundaries and differ from each other in terms of their impact on general functioning, relationships to life events, and responsiveness to different treatments options -- being able to monitor individual symptom trajectories over time is vital for effective treatment management \cite{fried_depression_2015,fried_moving_2017,cross_tracking_2015}. A desirable feature of intelligence-driven psychiatric support tools is that they directly quantify the contribution of different symptom severity estimates to overall condition risk in a modifiable way: in order to allow supervising clinicians (in dialogue with their clients) to maintain decision-making autonomy and preserve the therapeutic relationship from threats associated with fully automated or opaque assessment methods \cite{martin_estimating_2024}. In our implementation, capacity for providing accurate symptom-level readouts was supported by the use of a layer of surrogate models to compress rich feature input spaces to dimensions suitable for triangulation by a Bayesian network arbitrator -- a scalable method that is naturally extendable to other relevant sources of information (for example cognitive test features for concentration or motivational problems or passive-sensor data for somatic symptoms such as restlessness and sleep disturbance \cite{fara_bayesian_2023}). The underlying Bayesian network architecture intrinsically supports direct clinical intervention at the individual symptom level,  for example by isolating those better explained by contextual life factors than current mental wellbeing from network predictions.

Underwriting the potential success of intelligence-driven tools in healthcare is the availability of data with sufficient depth, breadth, and diversity to both maximize generalizability and support rigorous testing of important properties such as calibration accuracy and demographic fairness. Here, we were able to make use of one of largest to our knowledge consented and labelled mental health voice datasets (representing over 30,000 unique speakers, around a third of whom had a mental health diagnosis), to carry out robust development, calibration, and independent evaluation testing \cite{rutowski_toward_2024}. This degree of scale is likely to be necessary for satisfying recent best practice guidelines for health-related prediction models \cite{de_hond_guidelines_2022} as well as emerging regulatory standards for AI-based tools (for example, the recent ISO/IEC 42001 framework for AI governance \cite{ISO42001:2023}). 

Other strengths of our results include resilience of overall condition predictions to heterogeneity in symptom presentations, and demonstrated redundancy across different speech and voice input modalities. This is an important feature of multimodal models: as evidence of relevant signal across different input types (here, paralinguistic \textit{vs} linguistic feature predictions) reduces exposure to potential fragilities in any individual data source (for example, high background noise for acoustic features, or unwillingness to disclose sensitive information, insight problems, or cross-cultural expression differences for linguistic features). Critically, in-depth demographic fairness assessment revealed good-to-excellent properties in terms of potential for outcome allocation bias across age, gender, and race/ethnicity groups: with some evidence that demographic-specific calibration strategies may be useful for mitigating sex-based differences for anxiety. Performance was also found to be well-matched for individuals with and without comorbid chronic health conditions which are likely to impact voice (most commonly in our participants, asthma or other respiratory disorders and cardiovascular conditions; \autoref{tab:demogs}), and across recording device types (mobile devices \textit{vs} laptops). Finally, condition severity estimates (calculated as the sum of predicted severities across individual symptoms for each disorder) were found to be strongly associated with important patient-reported outcome measures (quality of life and psychosocial functioning scores) -- indicating validity of predictions in terms of symptom impact on functioning.

The model presented here also has some important limitations. In order to facilitate scale, datasets used self-reported information for current experience of depression and anxiety symptoms, as well as mental health diagnostic history. Future validation efforts incorporating and testing against clinician-provided diagnostic labels and symptom estimates would therefore be valuable. Further, additional testing of output robustness to a wider range of accent types, and in non-first language speakers, would be important prior to deployment in real-world clinical settings, where language familiarity and proficiency may be pre-existing barriers to access. Finally, it is vital that the introduction of new digital support tools occurs within the context of repeated acceptability and usefulness testing with the people who will use them, which here includes both healthcare practitioners and mental health service users \cite{timmons_bridging_2025}. Whilst both existing literature and our service user survey results revealed that, for some people, new assessment tools may help address perceived shortcomings in current screening and monitoring practices -- implementations should be mindful of user requirements for trustworthy systems. For example, whilst time to complete is a factor affecting the acceptability of standard monitoring measures (particularly if it is felt to detract from within-appointment care time \cite{bendall_contending_2020,faija_using_2022}), our survey results indicated that for speech and voice-based tools, concerns that limiting sample recording time might restrict expression may outweigh those around time-efficiency. Overall, assurances as to data security, privacy, and use-control, augmentation rather than displacement of clinical contact time, and maintenance of alternative monitoring options have been identified as key factors governing the acceptability of new assessment technologies in mental health services \cite{martinez-martin_ethical_2021,nice_digital_2025}.

In conclusion, we have argued that Bayesian network models offer a principled, explainable, and intervenable method for multi-source information integration that can be used to support clinical decision-making during assessment for common mental health problems. Our implementation demonstrates that with sufficiently large sample sizes, these models can achieve robust performance whilst maintaining transparent probabilistic reasoning that allows clinicians to both inspect dependencies between symptoms and conditions and incorporate their own insights into individual case structure. Appropriate scale also enables comprehensive fairness testing for different demographic groups -- which in combination with other model properties may help address several key previously-identified barriers to implementing digital phenotyping methods in the clinic \cite{huckvale_toward_2019}. Bridging the gap between traditional psychiatric assessment and data-driven approaches while preserving clinical autonomy represents a viable path towards precision psychiatry that enhances rather than replaces expert clinical judgment \cite{kabrel_current_2025}.

\section*{Methods}
\label{sec:methods}

\subsection*{Ethical approval and informed consent}
The ethical review process for this study was led by an independent research ethics expert (Dr. David Carpenter), on behalf of the Association of Research Managers and Administrators (ARMA, https://arma.ac.uk/), with a favourable opinion granted on September 2, 2024. All participants gave written informed consent and were compensated for their time. All study procedures were performed in accordance with the Declaration of Helsinki.

\subsection*{Data collection}
Participants were recruited using an online research participation platform (Prolific) and required to be 18 years or older, speak English as a first language, and resident in the US or UK. During recruitment, quotas were applied to ensure sufficient sampling of individuals of experience with significant mental health challenges (previous condition diagnosis), and at least nationally representative levels of race/ethnicity diversity.

Speech activity and self-report data were collected using the thymia research platform \cite{fara_speech_2022}. Participants completed two types of speech activity, both around 1 minute in length:
\begin{enumerate}[itemsep=0pt]
    \item \textbf{Reading out loud task}. Participant read out loud a standard text commonly used as a speech elicitation task due to its phonetic range (the Aesop fable "The North Wind and the Sun" \cite{deterding2006north}).
    \item \textbf{Answering a question task}. Participants were asked to describe what their mood had been like over the past two weeks, speaking at their usual volume and pace of speech.
\end{enumerate}

Self-report measures included validated questionnaire measures of depression and anxiety symptoms (the PHQ-8\cite{kroenke_phq-8_2009} and GAD-7\cite{spitzer_brief_2006}), mental health history (including past and current psychiatric diagnoses), and demographic information.  

\subsection*{Dataset construction}

For corpus construction, speech activity recordings were passed to an in-house pre-processing pipeline that involved resampling to a 16kHz rate and transcript generation using automated speech-to-text tools (Deepgram). For acoustic feature generation, silences were trimmed to produce audio segments of durations in the range $10-20s$. Segments without sufficient speech to reach this threshold, or where recording quality was not sufficient to support automated transcript generation, were excluded from analysis. To balance data quality against model generalizability to real-world settings, minimal data cleaning was applied. Specifically, we further excluded samples from participants with very short completion times for PHQ-8 or GAD-7 measures ($<1.5s$ per item, a simple indicator of potential inattentive responding\cite{ulitzsch_accounting_2024}) and/or mean speech-to-text transcription confidence of $<0.80$ ($\sim10\%$ of processed speech data).

Following these steps, a development dataset was constructed, consisting of 39,571 total speech activities from \textit{N}=21,379 users (see \autoref{tab:demogs}). Two further datasets, a calibration set (11,334 observed speech activities from \textit{N}=6,325 users) and held-out test set (4,866 observations from \textit{N}=2,431 users) were drawn from data collected after the development set was constructed and model structure was determined. No users overlapped between datasets, and the held-out test set was completely unseen during model development.

\subsection*{Analysis}

\subsubsection*{Surrogate models}

Since it would be computationally intractable to train a Bayesian network with the high-dimensional feature arrays used to represent speech acoustic and linguistic features, surrogate models were used to compress features into efficient representations of the signal contained in each feature type for each network symptom. Maintaining a level of unimodality at the surrogate representation level allowed delegation of integration of these different signals for each symptom to the Bayesian Network (see below). 

To maximise signal strength, surrogate models were trained on binary classification targets derived from item-level PHQ-8 and GAD-7 response data. Individual symptom severity scores were binarized to categories representing significant symptom presence (experienced on half or more days over the past 2 weeks) \textit{vs} mild or absent levels (experienced on less than half of days) -- in line with previously-established algorithms for converting PHQ data to DSM-like symptom presence criteria \cite{kroenke_phq-9_2001}. For each symptom, three types of surrogate models were trained and evaluated (45 models total):
\begin{enumerate}[itemsep=0pt]
    \item \texttt{reading-MM} models. Symptom predictions from combined acoustic and paralinguistic natural language processing (NLP) features from reading task data. Acoustic features were extracted from trimmed audio segments using an in-house speech model based on \texttt{TRILLsson5} \cite{shor_trillsson_2022}. Paralinguistic NLP features included speech rate, time to first utterance, and pause durations.
    \item \texttt{mood-audio} models. Symptom predictions from the same acoustic features extracted from trimmed mood question activity audio segments.
    \item \texttt{mood-linguistic} models. Symptom predictions from a linguistic feature set extracted from full mood question activity transcripts. Linguistic features were semantic embeddings from \texttt{ModernBERT-base} \cite{modernbert} plus an extended set of NLP features describing parts of speech such as first person pronoun use, generated using \texttt{spaCy} \cite{Honnibal_spaCy_Industrial-strength_Natural_2020}.
\end{enumerate}

\begin{figure}[H]
    \makebox[\textwidth][c]
    {\includegraphics[width=0.95\textwidth]{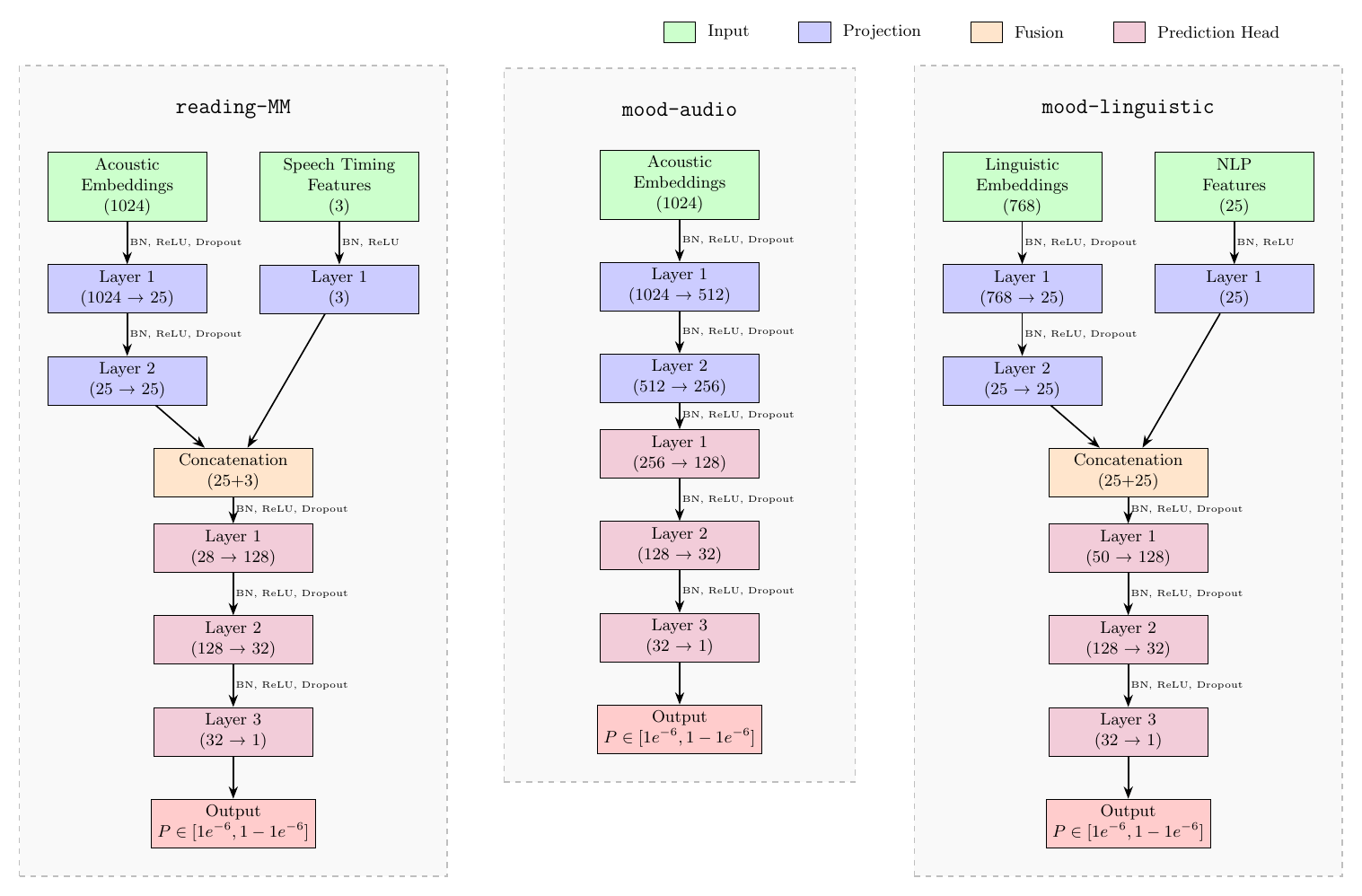}}
    \caption{\textbf{Surrogate model architecture}. Architecture of the three surrogate model types. All surrogate models were feedforward neural networks. \textit{BN}, batch normalization.} \label{fig:surrogate-architecture}
\end{figure}

\textbf{Model architecture}. Surrogate models were feedforward multilayer neural networks (\autoref{fig:surrogate-architecture}). 
\begin{itemize}[itemsep=0pt, topsep=3pt]
    \item For \texttt{reading-MM} models, acoustic embeddings and paralinguistic NLP (speech timing) features were initially processed separately. Acoustic embeddings were projected into a common projection dimension space using a two-layer network with batch normalization, dropout, and ReLU activation functions between layers (1024 → 25 → 25; where 25 is a common projection dimension across models based on the length of the full NLP feature set). The smaller speech timing feature set was projected using a simpler single-layer network that preserved dimensionality, with batch normalization and ReLU. The two projected representations were then concatenated in an additive fusion step.
    \item \texttt{mood-audio} models were two-layer networks with batch normalization, ReLU activation, and dropout between layers (1024 → 512 → 256). 
    \item For \texttt{mood-linguistic} models, linguistic embeddings and traditional NLP features were initially processed separately. Linguistic embeddings were projected using a two-layer network with batch normalization, ReLU, and dropout (768 → 25 → 25). NLP features were projected using a single layer network that preserved dimensionality, with batch normalization and ReLU. The two projected representations were then concatenated in an additive fusion step.  
\end{itemize}
For all surrogate model types, projections were passed to a prediction head that was a 3-layer network with a single output dimension, with batch normalization, ReLU, and dropout between each layer (combined projection dimension → 128 → 32 → 1). Final outputs were sigmoid-activated probabilities clamped to [0.000001, 0.999999].

\textbf{Hyperparameter tuning and training}. Surrogate models were trained using a nested cross-validation procedure in the development dataset, which was partitioned into training-validation and test splits in a 80:20 ratio. 

Model hyperparameters were tuned via stratified K-fold cross-validation in the training-validation split using Bayesian optimization via \texttt{Optuna} \cite{optuna} (k=4, with 50 Optuna trials per fold). Specifically, mean ROC-AUC across inner folds was optimized via tuning of learning rate, batch size, dropout rate, and weight decay.

Final models were then trained using optimal hyperparameter configurations in all development training-validation split data. Training used an AdamW optimizer with weight decay for regularization, binary cross-entropy loss, learning rate scheduling via ReduceLROnPlateau, early stopping based on validation loss and gradient clipping to mitigate against exploding gradients. Models were selected based on best ROC-AUC in validation data, and performance was then evaluated in the development test split.

\subsubsection*{Bayesian Network modelling}

The Bayesian network described here consists of nodes representing overall condition status for depression and anxiety, and nodes representing individual depression and anxiety symptom severity states. The network takes the form of a directed acyclic graph, in which condition probabilities influence symptom severity states, and symptom states may influence both other symptoms levels and the values of observable nodes (here: different speech-feature derived surrogate model predictions for each symptom). During training, we parameterize the network by providing it with sets of observed values for all network nodes (conditions, symptoms, and surrogate predictions), from which it learns conditional probability distributions for relationships between different nodes states (for example, the distribution across restlessness symptom severity categories, given different levels of psychomotor issues and trouble relaxing). At inference, we can query the network with evidence derived from speech activity data (surrogate model outputs), to generate predictions for both symptom and overall condition states.

\textbf{Network structure}. Given purely data-driven approaches to specifying network structure are known to be fragile (particularly for networks with larger numbers of nodes), we used a hybrid literature and data-informed approach to structure specification. 

Specifically, for specifying the inter-symptom network component, we leaned upon previously published large-sample joint network modelling analyses of depression and anxiety symptoms (predominantly measured using PHQ and GAD scales; \cite{beard_network_2016,kaiser_unraveling_2021,hoffart_network_2021,cai_network_2024}). From these studies, the most commonly identified and strongest intra-condition symptom connections (representing undirected partial correlations across participants at the same measurement occasion) for depression were between:
\begin{itemize}[itemsep=0pt, topsep=3pt]
    \item Low Mood and Anhedonia
    \item Sleep Issues and Low Energy
    \item Low Mood and Worthlessness Issues
    \item Psychomotor Issues and Concentration Issues
    \item Anhedonia and Low Energy
    \item Low Energy and Appetite Issues
    \item Low Mood and Low Energy
\end{itemize}

And for anxiety were between:
\begin{itemize}[itemsep=0pt, topsep=3pt]
    \item Excessive Worry and Uncontrollable Worry
    \item Trouble Relaxing and Restlessness
    \item Uncontrollable Worry and Nervousness
    \item Nervousness and Trouble Relaxing
    \item Nervousness and Dread
\end{itemize}

The most commonly identified between-condition symptom associations ("bridge symptoms") were between psychomotor symptoms and  restlessness, and concentration issues and trouble relaxing.

Inferring within-participant directed effects between symptoms requires intensive longitudinal data. Where possible, inter-symptom edge directions were informed by results temporal network analyses, which give insight into Granger-causal relationships between symptoms (i.e., whether increases or decreases in symptom A tends to precede increases or decreases in symptom B, between successive study time-points). Specifically, increases in self-reported low energy have been seen to precede increases in anhedonic symptoms \cite{ebrahimi_within-_2021,lunansky_disentangling_2023}, and worthlessness issues have been observed to precede increases in low mood \cite{ebrahimi_within-_2021}.

Key differences between these analyses and our setting are that 1) in order to perform inference on conditions as well as symptoms, the network contains direct mappings from constituent symptoms to condition nodes and 2) that symptom estimates are informed by observable inputs (surrogate model predictions) that, although separately trained for each symptom, together hold information about the correlation structure across symptoms (for example, participants with higher excessive worry predictions will also tend to have higher uncontrollable worry predictions, since these are correlated at the ground truth level). During development testing it was found that indirect association via common association with parent depression or anxiety nodes and relationships between surrogate model predictions was sufficient to preserve many of the literature-identified inter-symptom associations in model predictions, without the need for specifying direct inter-symptom edges in the network. The retained set of sufficient inter-symptom network edges were:
\begin{itemize}[itemsep=0pt, topsep=3pt]
    \item Low Energy → Anhedonia
    \item Worthlessness Issues → Low Mood
    \item Appetite Issues → Low Energy
    \item Trouble Relaxing → Restlessness
    \item Psychomotor Issues → Restlessness
    \item Trouble Relaxing → Concentration Issues
\end{itemize}

To define surrogate prediction to symptom node edges, we started with a fully connected network (with all surrogates informing all relevant symptom estimates), then pruned connections according to relative informativeness (according to both overall performance strength and cross-symptom specificity) -- with the constraint that each symptom must be informed by at least one two surrogate observations, with one being an acoustic features model (i.e., no symptoms were informed only by linguistic features). This process was intended to encourage the network to make use of symptom-specific information, rather than leaning on shared variance across symptoms. Effects of pruning explicit inter-symptom and symptom-observable edges on model performance and symptom prediction covariance were evaluated in the development set test split, with optimal network configuration chosen to optimize symptom and condition level performance (ROC-AUC) whilst retaining cross-symptom prediction covariance similar to the ground truth correlation structure.

\begin{figure}[H]
    \makebox[\textwidth][c]
    {\includegraphics[width=0.60\textwidth]{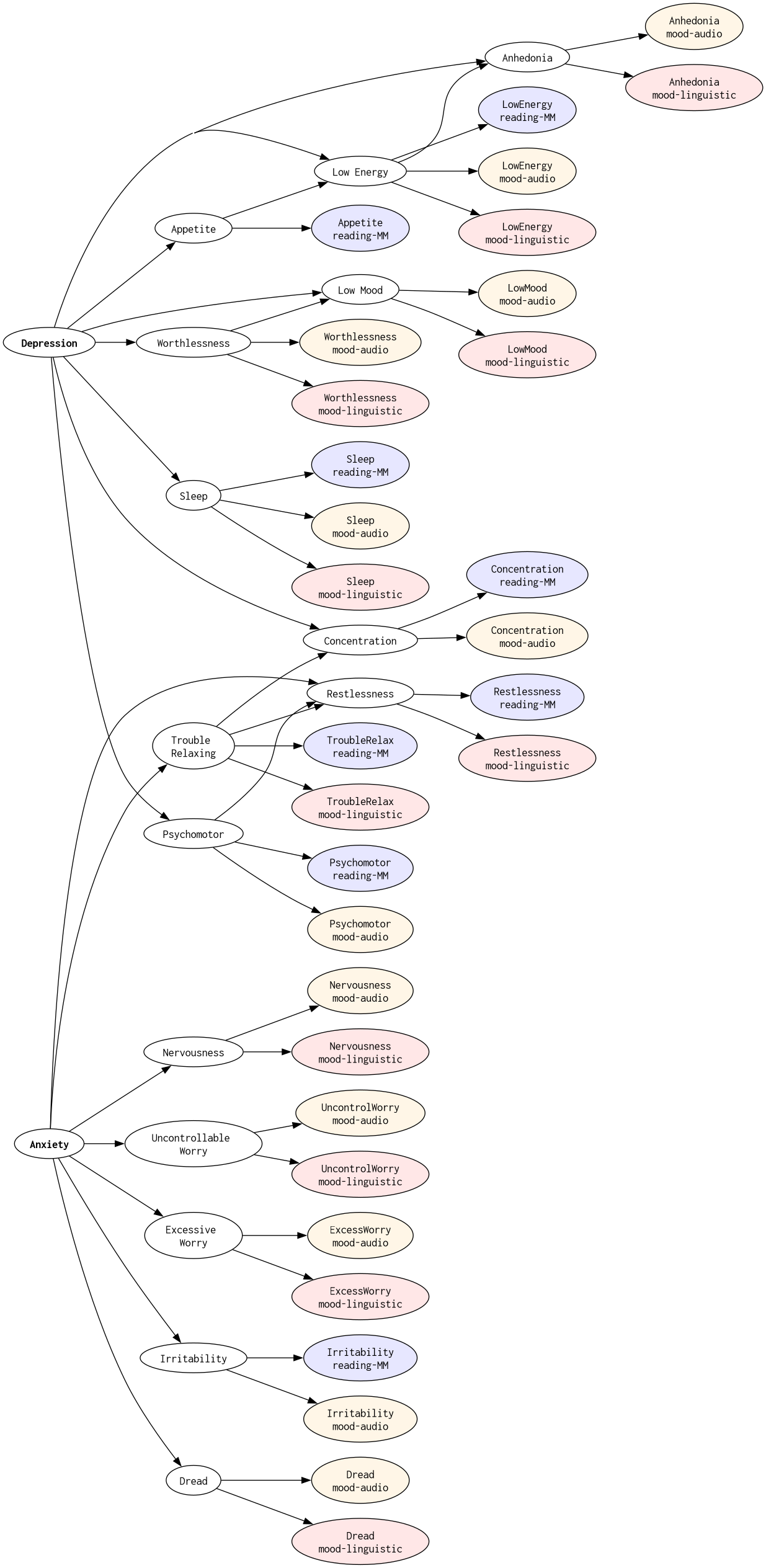}}
    \caption{\textbf{Bayesian Network structure}. A joint network of both overall depression and anxiety status, and individual symptom severity states was constructed. Arrows represent the direction of causality implied by the specified direct edges (connections between nodes). Filled nodes represent observed evidence used at inference.} \label{fig:bn-structure}
\end{figure}

For a diagram of the final network structure, including all retained symptom-surrogate edge connections, see \autoref{fig:bn-structure}. 

In order to support efficient estimation and exact inference methods, the model was specified as a Discrete Bayesian Network, using \texttt{pgmpy} \cite{Ankan2024}. 

\textbf{Data processing and target definitions}. Ground truth values for symptom severity levels already existed in the form of discrete categories (PHQ-8 and GAD-7 item scores). Surrogate model predictions were continuous probabilities for symptom presence and were therefore discretized prior to entering the network using quartile transforms (with cut boundaries defined from observed distribution of each model's predictions in the development set).

\begin{figure}[H]
    \makebox[\textwidth][c]
    {\includegraphics[width=0.5\textwidth]{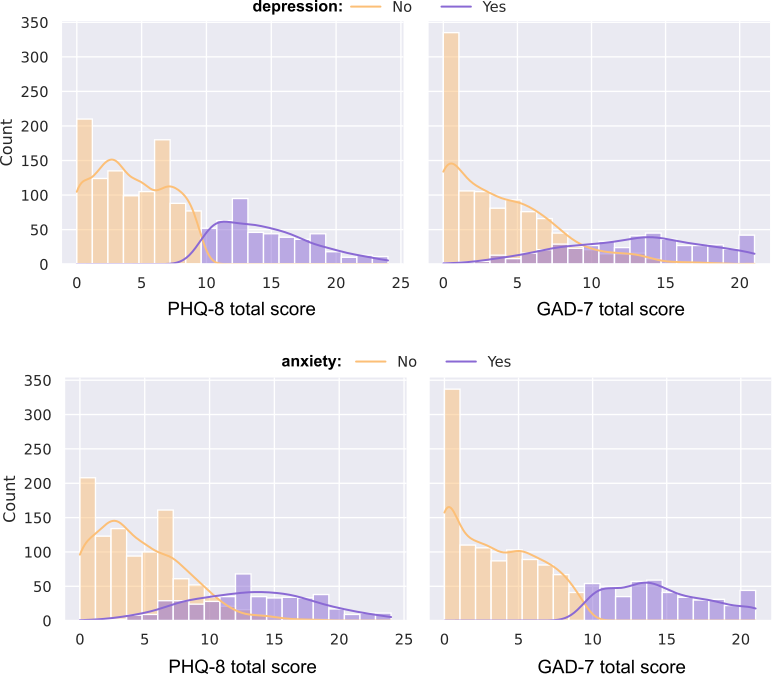}}
    \caption{\textbf{Distribution of PHQ-8 and GAD-7 total scores in test set data, by condition}.} \label{fig:scores-by-target}
\end{figure}

For condition status, we used a compound target definition based on established cut-off scores for PHQ-8 and GAD-7 measures \cite{kroenke_phq-8_2009,spitzer_brief_2006} in conjunction with self-reported mental health condition diagnosis information. Specifically, we defined depression/anxiety as being present when current PHQ/GAD total scores were $\ge10$ and a participant reported a previous mental health condition diagnosis, and absent when PHQ/GAD total scores were $<10$ and a participant reported no mental health condition diagnosis. This definition was chosen on the basis of previous experiments in our datasets that have shown improved performance for simple acoustic speech models when training on this definition compared to simple binary split approaches: whether testing on either these or binary split target definitions. This implies that there is an advantage at training to using target definitions which (indirectly) incorporate external clinical judgement. We verified that use of this definition does not artificially decrease problem difficulty relative to binary split approaches by confirming that target categories retain good coverage across possible PHQ-8 and GAD-7 severity ranges (see \autoref{fig:scores-by-target}). Additionally, we assessed performance robustness to alternative condition definitions which did not require presence or absence of a self-reported diagnosis (see Results).

\textbf{Parameter estimation}. Network parameters (posterior conditional probability distributions between state values) were estimated from development set training-validation split observations of ground truth and surrogate model prediction values using Bayesian estimation methods. 

Specifically, we used Bayesian Dirichlet equivalent uniform priors, which act to regularize the posterior conditional probability distributions, and allow for some posterior weight to remain on combinations of state values which are rare or unseen in training data (guarding against over-fitting). This form of prior is specified as pseudo counts which are equivalent to having observed \textit{N} uniform samples of each network variable state and parent configuration before the estimation data is observed (with observed data added to the pseudo count priors before normalization to produce posterior values). In \texttt{pgmpy}, equivalent uniform priors are specified by providing an equivalent sample size parameter, with the prior pseudo counts added to each node state combination calculated as $equivalent sample size/(node cardinality*product(parent cardinalities))$. 

During network development, we explored the effects of different degrees of prior regularization strength (equivalent sample size values) on ROC-AUC performance and correlation between output predictions for overall condition probabilities. This was motivated by the fact that depression and anxiety status were strongly correlated across individuals in our network training data, but a desired model property for generalization purposes is to ensure condition-specific sensitivity in predictions (allowing for cases where anxiety is present but not depression, and vice versa). Tested equivalent sample size parameter values effectively explored adding an additional 5-10\% prior counts to single condition presence configurations, with final models using $equivalent sample size = 8000$ (for our network specification and estimation dataset size this is equivalent to adding 250 or 7.5\% additional observations to each possible depression/anxiety node state configuration).

\textbf{Inference procedure}. Inference was performed by querying the parameterized model with unseen test data surrogate model predictions, using Variable Elimination (an exact inference method \cite{zhang_exploiting_1996}). 

\subsubsection*{Calibration}

Given the importance of accurate probability estimates to clinical usefulness, we used an additional data set (matched to the development set in terms of stratification for age, birth sex, race/ethnicity, and mental health diagnosis status) to train calibrator models for refining overall depression and anxiety predictions. Specifically, we trained a collection ($N=10$) of bagged \texttt{sklearn} \texttt{IsotonicRegression} estimators. Initially, 5-fold cross-validation was performed within the calibration set, in order to assess the stability of both standard calibration metrics (ECE, MCE, and Brier score) and discriminative performance (ROC-AUC) in calibrated predictions. With stable performance achieved, a final collection of calibrator models was trained using the entire calibration set.

\subsubsection*{Evaluation}

Final performance estimates are based on results obtained in the held-out test set (data unseen during model development), with results from development testing provided for reference. In order to provide a fair assessment of the network's multimodal capabilities, only samples from users with surrogate model predictions from all three model types were included. For conditions, performance is reported for all users where ground truth depression/anxiety status was defined (for each condition separately). For symptoms, performance is reported for all test data (for the exact \textit{N} included in each analysis see Results).

Following recent reporting recommendations for predictive AI models in medicine \cite{calster_performance_2024}, model performance for conditions is described using both discrimination (ROC-AUC) and calibration (ECE) metrics (using \textit{M}=10 bins), with full calibration plots provided. These specific metrics are recommended based on assessment of statistical properness, and whether or not they incorporate factors related to misclassification cost that are more appropriately dealt with by formal decision analysis and clinical utility calculations. For example, whilst it is sometimes argued that F1 and related metrics should be provided in cases of class imbalance, the guidelines state that these are inappropriate for clinical predictive models since these both ignore true negative rates (which are important for medical applications) and conflate classification performance with clinical utility. This is because the extent to which an outcome is imbalanced (an epidemiological feature of the data) is not mathematically proportional to the extent to which misclassification costs are imbalanced (a clinical characteristic related to the specific medical decisions supported by the model).

For fairness assessment, we report within-group discriminative performance, and group differences in ECE, Brier scores (the mean squared distance between observed and predicted outcomes, or degree of alignment between predicted and observed outcome probabilities), and equalized odds ratios (as implemented in \texttt{fairlearn} \cite{weerts2023fairlearn}), which assesses whether different group members experience differential outcome allocation in terms of true and false positive rates.

We additionally report prevalence-dependent performance metrics for our test population, calculated using an example threshold of 0.5 for calibrated condition probabilities. Positive predictive values represent the proportion of positive predictions that are true positives, and negative predictive values the proportion of negative predictions that are true negatives. Positive and negative likelihood ratios describe the change in odds of having a condition after receiving a positive and negative test result, respectively, for a given population.

\subsection*{Service user consultation}
For the service user consultation survey, we recruited a sample of \textit{N}=230 individuals from Prolific, with the following eligibility criteria: (1) UK residence; (2) aged 18 years or older; (3) self-reported current diagnosis of a mental health condition; (4) attempted to access NHS mental health support within the past 12 months; and (5) currently receiving or awaiting treatment (psychological therapy, medication, brain-stimulation treatments, hospital/residential programs, or self-management). Quota sampling ensured equal representation by birth sex and age ($\leq$40 vs $>$40 years).  

Participants provided their views about both standardized questionnaire measures and a new hypothetical voice-based mental health screening tool using likert ratings and open-ended free text responses. The voice-based mental health screening tool was described as a computer program to which short speech samples could be provided, that can take into account both the content of their words and other speech qualities (like tone, pace, or pauses) to provide estimates of their current mental health status to their healthcare provider ahead of an appointment.

Rating scale responses were summarised using medians and interquartile ranges. Free-text responses were analysed using thematic analysis adapted from Braun and Clarke \cite{braun_using_2006} with additional systematic coding procedures \cite{boyatzis1998transforming}. The first author (AN) familiarised herself with responses and inductively developed a coding framework capturing distinct patterns in participant feedback. Each response was coded for presence/absence of themes, with multiple themes possible per response. Recognising the applied nature of this research and need for reproducibility, we incorporated inter-coder reliability testing with a second independent coder (AG). Cohen's Kappa indicated substantial agreement across all categories (mean $\kappa$ = 0.83). Theme prevalence was then determined through a frequency count to identify the most salient concerns among participants.

\subsection*{Code availability}

Supporting analysis code is available to reviewers and editors on request during submission. 

\bibliography{bib}


\section*{Funding}
This research did not receive any external funding.

\section*{Author contributions statement}
S.G. and E.M. conceived the experiment(s),  A.N. and G.F. conducted the experiment(s), A.N., G.F. and A.L.G. analysed the results.  All authors reviewed and contributed to the writing of the manuscript. 

\section*{Data availability statement}

The datasets generated during and analysed during the current study are not publicly available due to lack of consent for public sharing of raw data, which due to the nature of speech data would compromise participant privacy. They may be made available on reasonable request to the corresponding author for non-commercial research purposes related to detecting or monitoring mental health and wellbeing (the purposes for which consent for research data sharing was obtained), subject to completion of a Data Sharing Agreement with thymia Limited.

\section*{Additional information}
\textbf{Competing interests}.

E.M. and S.G. are co-founders of thymia Ltd. A.N., G.F., and A.L.G. are employees of thymia Ltd. E.M., S.G., A.N., A.L.G. hold equity in the company, which may benefit from commercialization of technologies similar to those described in this paper. M.M.N. has acted as a paid consultant for thymia Ltd, which does not extend to involvement in the current work. 



\newpage

\supplementarymaterial
Supplementary Material for \textit{A multimodal Bayesian Network for symptom-level depression and anxiety prediction from voice and speech data} by Agnes Norbury, George Fairs, Alexandra L. Georgescu, Matthew M. Nour, Emilia
Molimpakis, and Stefano Goria (2025).

\newpage 

\begin{table}[h]
\centering
\begin{tabular}{@{}rrccc@{}}
\toprule
\multicolumn{1}{l}{} & \multicolumn{1}{l}{} & \begin{tabular}[c]{@{}c@{}}Development \\ (\textit{N}=21379)\end{tabular} & \begin{tabular}[c]{@{}c@{}}Calibration \\ (\textit{N}=6325)\end{tabular} & \begin{tabular}[c]{@{}c@{}}Test \\ (\textit{N}=2431)\end{tabular} \\ \midrule
Age & Mean (SD) & 37.6 (12.9) & 37.4 (13.2) & 37.1 (13.0) \\
\multicolumn{1}{l}{} & Range & 18-89 & 18-88 & 18-80 \\
Birth Sex & Female & 13031 & 4143 & 1637 \\
 & Male & 8290 & 2015 & 766 \\
 & Prefer not to say & 44 & 31 & 7 \\
 & Intersex & 14 & 7 & 3 \\
Gender & Woman & 12780 & 4091 & 1591 \\
 & Man & 8250 & 1988 & 759 \\
 & Non-binary & 303 & 93 & 56 \\
 & Prefer not to say & 46 & 24 & 7 \\
Race/Ethnicity & White & 15544 & 4508 & 1771 \\
 & Black & 3289 & 774 & 302 \\
 & Mixed & 1076 & 388 & 150 \\
 & Asian & 1054 & 385 & 141 \\
 & Other & 416 & 141 & 49 \\
Diagnosed Mental Health Condition & No & 12387 & 3466 & 1307 \\
 & Yes & 8597 & 2615 & 1057 \\
 & Prefer not to say & 259 & 115 & 49 \\
Mental Health Diagnosis & Depressive Disorder & 5718 & 1709 & 698 \\
 & Anxiety Disorder & 6460 & 1981 & 844 \\
 & Other & 1471 & 407 & 146 \\
PHQ-8 Total Score & Mean (SD) & 7.9 (5.7) & 7.6 (5.4) & 7.7 (5.4) \\
\multicolumn{1}{l}{} & Range & 0-24 & 0-24 & 0-24 \\
GAD-7 Total Score & Mean (SD) & 7.3 (5.7) & 7.1 (5.6) & 7.2 (5.6) \\
\multicolumn{1}{l}{} & Range & 0-21 & 0-21 & 0-21 \\
Chronic Physical Health Condition & None & 12301 & 3663 & 1391 \\
 & Respiratory & 4933 & 745 & 343 \\
 & Cardiovascular & 3419 & 956 & 373 \\
 & Arthritis & 1516 & 367 & 155 \\
 & Anaemia & 1257 & 357 & 153 \\
 & Sleep Apnoea & 1215 & 332 & 139 \\
 & Diabetes & 1107 & 299 & 114 \\
 & Fibromyalgia & 637 & 149 & 71 \\
 & Long COVID & 625 & 137 & 55 \\
 & Chronic Fatigue & 605 & 148 & 66 \\
 & Other & 2445 & 659 & 290 \\
Accent & US/Other & 11565 & 3668 & 1444 \\
 & UK & 9578 & 2488 & 957 \\
Device Type & Laptop & 12857 & 3567 & 1422 \\
 & Smartphone or Tablet & 8272 & 2605 & 984 \\ \bottomrule
\end{tabular}
\caption{\textbf{Participant numbers and demographics by study dataset.} Values represent \textit{N} unless otherwise specified.}
\label{tab:demogs}
\end{table}

\begin{table}[h]
\centering
\begin{tabular}{@{}rrcc@{}}
\toprule
Condition & Symptom & Model & ROC-AUC \\ \midrule
Depression & Anhedonia & mood-audio & 0.674 \\
 &  & mood-linguistic & 0.715 \\
 & Low Mood & mood-audio & 0.712 \\
 &  & mood-linguistic & 0.779 \\
 & Sleep & reading-MM & 0.620 \\
 &  & mood-audio & 0.662 \\
 &  & mood-linguistic & 0.684 \\
 & Low Energy & reading-MM & 0.634 \\
 &  & mood-audio & 0.692 \\
 &  & mood-linguistic & 0.724 \\
 & Appetite & reading-MM & 0.620 \\
 & Worthlessness & mood-audio & 0.691 \\
 &  & mood-linguistic & 0.746 \\
 & Concentration & reading-MM & 0.601 \\
 &  & mood-audio & 0.649 \\
 & Psychomotor & reading-MM & 0.638 \\
 &  & mood-audio & 0.680 \\
Anxiety & Nervousness & mood-audio & 0.709 \\ 
 &  & mood-linguistic & 0.742 \\
 & Uncontrollable Worry & mood-audio & 0.695 \\
 &  & mood-linguistic & 0.733 \\
 & Excessive Worry & mood-audio & 0.692 \\
 &  & mood-linguistic & 0.735 \\
 & Trouble Relaxing & reading-MM & 0.607 \\
 &  & mood-linguistic & 0.714 \\
 & Restlessness & reading-MM & 0.624 \\
 &  & mood-linguistic & 0.652 \\
 & Irritability & reading-MM & 0.623 \\
 &  & mood-audio & 0.677 \\
 & Dread & mood-audio & 0.654 \\ 
 &  & mood-linguistic & 0.682 \\ \bottomrule
\end{tabular}
\caption{\textbf{Discrimination performance for individual surrogate models.} Results are reported for the development set test split where surrogate model performance was evaluated, for all surrogates used as inputs to the final (pruned) Bayesian network model.}
\label{tab:surrogate-symptom-rocs}
\end{table}

\begin{table}[h]
\centering
\begin{tabular}{@{}rcccc@{}}
\toprule
 & \begin{tabular}[c]{@{}c@{}}reading-MM\\ only\end{tabular} & \begin{tabular}[c]{@{}c@{}}mood-audio \\ only\end{tabular} & \begin{tabular}[c]{@{}c@{}}mood-linguistic\\ only\end{tabular} & \begin{tabular}[c]{@{}c@{}}reading-MM \& \\ mood-audio\end{tabular} \\ \midrule
ROC-AUC Depression & 0.717 & 0.772 & 0.783 & 0.783 \\
ROC-AUC Anxiety & 0.705 & 0.776 & 0.805 & 0.783 \\
ECE Depression (Uncalibrated) & 0.052 & 0.060 & 0.091 & 0.078 \\
ECE Anxiety (Uncalibrated) & 0.077 & 0.084 & 0.086 & 0.077 \\ \midrule
ROC-AUC Anhedonia & 0.589* & 0.681 & 0.725 & 0.680+ \\
ROC-AUC Low Mood & 0.619* & 0.716 & 0.779 & 0.716+ \\
ROC-AUC Sleep & 0.627 & 0.669 & 0.693 & 0.674 \\
ROC-AUC Low Energy & 0.634 & 0.686 & 0.723 & 0.686 \\
ROC-AUC Appetite & 0.629 & 0.670* & 0.689* & 0.669+ \\
ROC-AUC Worthlessness & 0.627* & 0.702 & 0.751 & 0.707+ \\
ROC-AUC Concentration & 0.620 & 0.660 & 0.719* & 0.663 \\
ROC-AUC Psychomotor & 0.640 & 0.693 & 0.699* & 0.691 \\
ROC-AUC Nervousness & 0.618* & 0.708 & 0.746 & 0.706+ \\
ROC-AUC Uncontrollable Worry & 0.615* & 0.702 & 0.738 & 0.700+ \\
ROC-AUC Excessive Worry & 0.621* & 0.697 & 0.737 & 0.696+ \\
ROC-AUC Trouble Relaxing & 0.614 & 0.678* & 0.715 & 0.665+ \\
ROC-AUC Restlessness & 0.625 & 0.658* & 0.671 & 0.659+ \\
ROC-AUC Irritability & 0.631 & 0.689 & 0.725* & 0.689 \\
ROC-AUC Dread & 0.599* & 0.663 & 0.694 & 0.664+ \\ \bottomrule
\end{tabular}
\caption{\textbf{Performance for Bayesian Network when supplied only with different surrogate model prediction types}.Results are reported for the development set test split where effects of Bayesian network architecture on performance were explored. Given in the final (pruned) network structure not all symptoms are informed by each surrogate type, some symptoms in each single-surrogate analysis are informed only by connections with other symptom estimates (*). For the combined paralinguistic surrogates results (reading-MM \& mood-audio), + indicates symptoms only directly informed by one of the two paralinguistic surrogate types. Results are reported for the development set test split where effects of Bayesian Network structure on performance were evaluated.}
\label{tab:single-surrogate-bn-performance}
\end{table}

\begin{table}[h]
\centering
\begin{tabular}{@{}rll@{}}
\toprule
\multicolumn{2}{c}{Variable} & \multicolumn{1}{c}{Value} \\ \midrule
Age & Mean (SD) & 41.0 (12.0) \\
\multicolumn{1}{l}{} & Range & 19-69 \\
Birth Sex & Female & 115 \\
 & Male & 115 \\
Race/Ethnicity & White & 206 \\
 & Mixed & 8 \\
 & Black & 7 \\
 & Asian & 7 \\
 & Other & 2 \\
Mental Health Diagnosis & Anxiety & 179 \\
 & Depression & 172 \\
 & PTSD & 31 \\
 & OCD & 27 \\
 & Eating Disorder & 16 \\
 & Bipolar Disorder & 14 \\
 & Other & 38 \\
Treatment Experience & Medication & 145 \\
 & Psychological therapy & 101 \\
 & Own care & 66 \\
 & Hospital and residential & 6 \\
 & Other & 3 \\
Length of Experience & More than 10 years & 163 \\
 & 6-10 years & 33 \\
 & 3-5 years & 22 \\
 & 1-2 years & 8 \\
 & Less than 1 year & 3 \\
 & Prefer not to say & 1 \\
Questionnaire Familiarity & Very familiar & 105 \\
 & Somewhat familiar & 116 \\
 & Not very familiar & 9 \\
\multicolumn{1}{l}{Comfort with Technology} & Very comfortable & 158 \\ 
 & Somewhat comfortable & 58 \\
 & Moderately comfortable & 11 \\
 & Not very comfortable & 3 \\ \bottomrule
\end{tabular}
\caption{\textbf{Description of participants for the stakeholder consultation survey}. All participants were based in the UK, had a current mental health diagnosis, and reported that they had tried to access mental health support via the NHS in the past 12 months (total \textit{N}=230). Values represent N unless otherwise specified (diagnosis and treatment experience categories are non mutually-exclusive).}
\label{tab:survey-demogs}
\end{table}

\begin{table}[h]
\centering
\begin{tabular}{@{}ll@{}}
\toprule
Theme & \textit{N} Mentions \\ \midrule
\begin{tabular}[c]{@{}l@{}}Strong privacy policy; effective cybersecurity; demonstrated compliance with relevant data protection legislation: \\ including full information on who has access to the data and what purposes it will be used for\end{tabular} & 127 \\
\begin{tabular}[c]{@{}l@{}}Evidence of tool accuracy and reliability (e.g., published studies and clinical trials);\\ including evidence of fair performance for people from my specific demographic group \end{tabular} & 81 \\
\begin{tabular}[c]{@{}l@{}}Recommendation by care provider (e.g., GP) or health system (e.g., NHS)\end{tabular} & 18 \\
At this moment in time, nothing would make me trust or be comfortable with this kind of tool (AI scepticism) & 15 \\
Recommendations from other end-users (e.g., testimonials, reviews) & 14 \\
Option to test out and self-verify results (check if these align with my and/or my clinician's impressions) & 10 \\
Easy to use interface; well-tested, with technical support available if needed & 7 \\
\begin{tabular}[c]{@{}l@{}}Assurance that results will always be interpreted in conjunction with input from my clinician \\ (and not replace human clinical interactions)\end{tabular} & 5 \\
\begin{tabular}[c]{@{}l@{}}Ability to control which speech recordings are analysed or passed to clinicians; \\ option to re-record responses if I want \end{tabular} & 5 \\
Ethical implementation; safeguards against misuse; accountability for errors & 4 \\
Assurance that I will be able to see my results and challenge them if I don't agree & 4 \\ \bottomrule
\end{tabular}
\caption{ \textbf{Qualitative analysis of features required for trustworthy implementation of a speech-based mental health assessment tool according to a UK-based sample of mental health service users}. Thematic analysis of feedback provided by \textit{N}=230 respondents, alongside number of individual participants referencing each theme (for themes mentioned by 2 or more participants).}
\label{tab:voice-tool-trustworthiness}
\end{table}

\end{document}